# V. Kromer

# Mathematical Model of Word Length on the Basis of the Čebanov-Fucks Distribution with Uniform Parameter Distribution

1. There are several mathematical models of word length measured by the number of syllables comprising it. The simplest one is the Poisson distribution when the obligatory (first) syllable is not taken into account (the Čebanov-Fucks distribution). Its merit consists in the linguistic interpretation of the only parameter – the average word length in the text. The truncated Poisson distribution and a number of two-parameter distributions (negative binomial, Hyper-Poisson, etc.) are also applied.

2. The author of this paper earlier suggested a word length model, based on the Čebanov-Fucks distribution, with a random parameter, which is uniformly distributed within the interval $(\lambda_1, \lambda_2)$, where $\lambda_1$ and $\lambda_2$ are the parameters of the suggested distribution. The linguistic sense of the parameters $\lambda_1$ and $\lambda_2$ is mathematical expectation of the word length correspondingly at the beginning and at the end of the list of textual words, ranged according to the decrease of their occurrence. The uniform distribution of the parameter $\lambda$ of the Čebanov-Fucks distribution conforms with the principle of the optimal coding according to which word frequencies in the rank frequency list are distributed in conformity with the Estoup-Zipf-Mandelbrot law, and the mathematical expectation of the word length is linearly dependent on the logarithm of its rank. The parameters $\lambda_1$ and $\lambda_2$ are determined by minimizing the Chi-square criterion and by normalizing the total number of text words.

3. In accordance with the suggested model, the data on 13 typologically different languages, ancient and modern, have been processed. In some languages the data were computed on texts belonging to different genres (styles). The correspondence between the theoretical and empirical distributions is evaluated by the discrepancy coefficient $C = \dfrac{\sum \chi^2}{N}$, where $N$ is the sample size, measured in number of words, $\sum \chi^2$ is the sum "Chi-square". The correspondence is considered satisfactory at $C \leq 0.02$. In case $\lambda_1$ is equal to $\lambda_2$ the suggested distribution degenerates into the Čebanov-Fucks distribution. For some languages (Chinese, Hebrew, Icelandic, Turkish) the correspondence at this value is unsatisfactory. The suggested model is not fit for word structure description of the said languages. With $\lambda_1$ equal to $\lambda_2$ and a satisfactory correspondence the model is considered fit for the word structure description (Latin, Czech, Quechua). More interesting is the case when $\lambda_1$ is not equal to $\lambda_2$; the correspondence then is always satisfactory. The value $\lambda_0 = \dfrac{\lambda_1 + \lambda_2}{2}$ is the average word length in the text. To every parameterized text there corresponds a dot on the plane $\lambda_0 \lambda_1$. The analysis allows us to conclude that texts of the same genre are characterized by direct statistical dependence of $\lambda_1$ on $\lambda_0$. The dependence for the genre of letters proves to be correct in English, French, German, Swedish, Spanish and Italian (the languages are listed in the order of increasing the average word length). The analyzed material doesn't include Luther's letters. The Latin phrases occurring in them lead to sharp deviation of the word structure from the general tendency, in particular the $\lambda_1$ value is excessive (the coefficient of completion $\alpha$ is small, see p. 4; for Latin texts $\alpha = 0$). With the increase of $\lambda_0$ from 1.37 to 2.21 the parameter $\lambda_1$ changes from 0.73 to 1.19. The dependence is approximately linear and is approximated through the function $\lambda_1 = 0.18 + 0.45\lambda_0$. The reliability of approximation $R^2$ is equal to 0.79.

4. For the German language, more detailed and consistently represented in accessible data, there appears an opportunity to define precisely the $\lambda_1(\lambda_0)$ dependence for texts of various genres. With the change of $\lambda_0$ from 1.45 to 2.13 the $\lambda_1$ parameter changes from 1.33 to 0.81. The dependence is inverse and is satisfactorily approximated by the power function. Since the value $\lambda_0$ doesn't go lower than 1 owing to the condition that each word must have at least one syllable, and the $\lambda_1$ value asymptotically goes to its lowest limit $\lambda_{1\,min}$ (at the given stage of research we consider it equal to 0.5), the dependence to be found is approximated by the power function with the displaced origin of coordinates: $\lambda_1 = 0.5 + \dfrac{0.34}{(\lambda_0 - 1)^{1,01}}$. The approximation reliability equals 0.95. The increase of the range $(\lambda_2 - \lambda_1)$ with the $\lambda_0$ increase allows us to conclude that the given dependence reflects the degree of completion of synergetic processes of code (language) optimization. For more "exquisite" genres the completion degree must be higher than for simple genres. Let's assume that for a completely optimized language the $\lambda$ parameter in the Čebanov-Fucks distribution changes from $\lambda_{1\,min}$ to some maximum value. With an incomplete optimization, characterized by the completion coefficient $\alpha$ $(0 \leq \alpha \leq 1)$, the parameter $\lambda_1$ value will make $\lambda_1 = \alpha\,\lambda_{1\,min} + (1 - \alpha)\lambda_0$, there follows $\alpha = \dfrac{\lambda_0 - \lambda_1}{\lambda_0 - \lambda_{1\,min}}$. With $\lambda_{1\,min} = 0.5$ the $\alpha$ coefficient consistently, as $\lambda_0$ increases and $\lambda_1$ decreases, acquires the following values: 0.12 for Middle High German; 0.20 for the Baroque verses; 0.56 for letters of the writer K. Tucholsky; 0.66 for letters of the 19[th] and the first half of the 20[th] centuries; 0.77 for Scientific texts and 0.81 for newspaper texts. Table 1 shows the languages and genres with corresponding dots. Fig. 1 shows the $\lambda_1(\lambda_0)$ dependence for 12 texts and the approximating curves.

5. For named in p. 3 six European languages (the genre of letters) the average value of the completion coefficient makes 0.63 with the standard deviation 0.06 and doesn't reveal any significant dependence on $\lambda_0$. Fig. 2 shows the $\alpha(\lambda_0)$ dependence for the genre of letters. The horizontal line is the average $\alpha$ value.

Table 1

| Dot number | Language | Genre |
|---|---|---|
| 1 | English | Letters |
| 2 | French | Letters |
| 3 | German | Kurt Tycholsky's letters |
| 4 | Swedish | Gunnar Ekelöf's letters |
| 5 | German | Letters of the 19th and the first half of the 20th centuries |
| 6 | Spanish | F.G. Lorca's letters |
| 7 | Spanish | Gabriela Mistral's letters |
| 8 | Italian | Letters |
| 9 | Middle High German | |
| 10 | German | Baroque verses |
| 11 | German | Scientific texts |
| 12 | German | Newspaper texts |

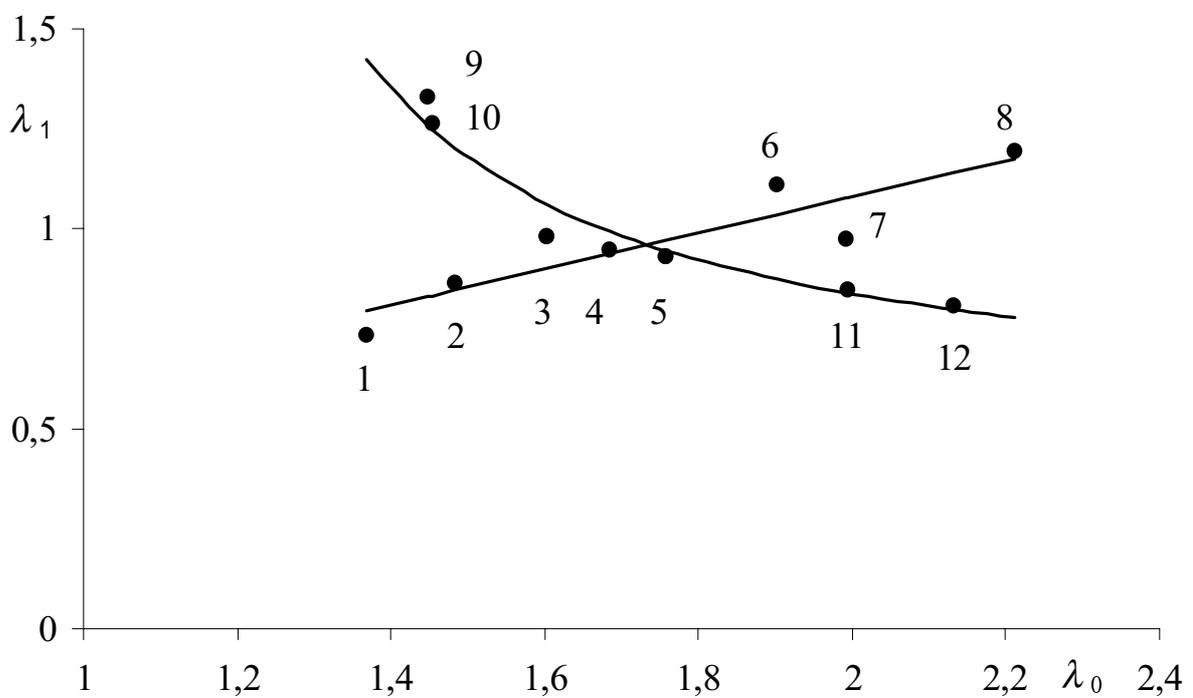

Fig. 1.

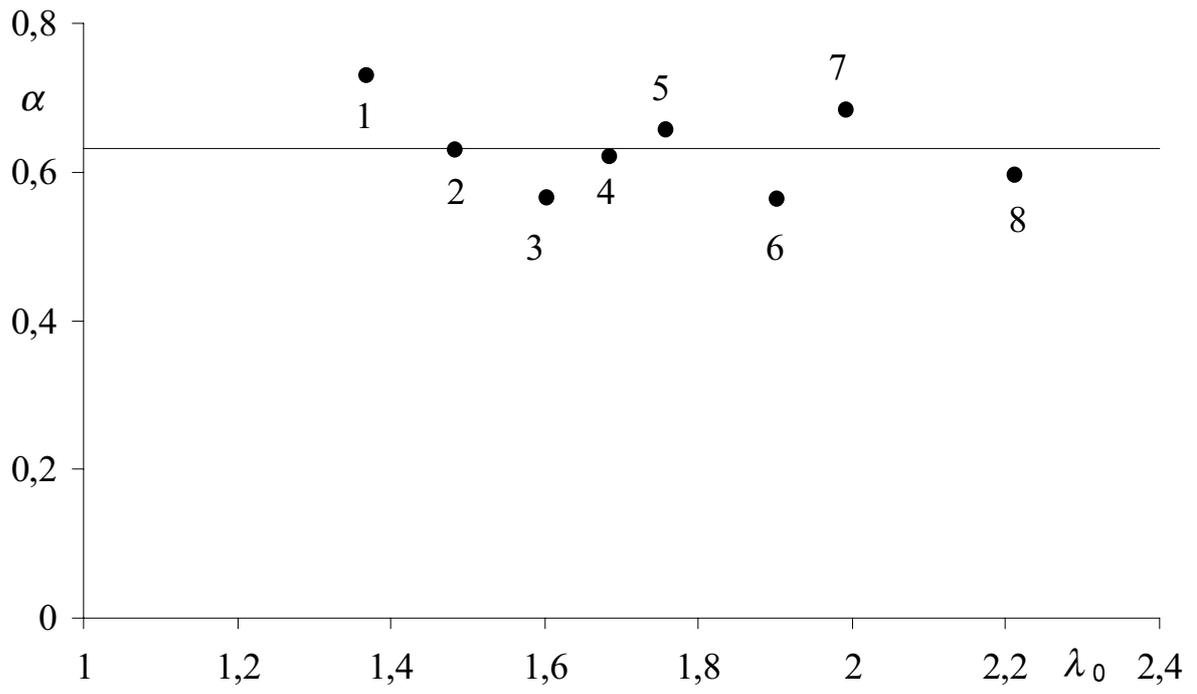

Fig. 2.